\documentclass{article}
\usepackage{spconf,amsmath,graphicx}
\usepackage{amssymb, multirow, booktabs}

\title{LIGHTWEIGHT AND EFFICIENT END-TO-END SPEECH RECOGNITION\\ USING LOW-RANK TRANSFORMER}
\name{Genta Indra Winata$^\star$ \qquad Samuel Cahyawijaya$^\star$ \qquad Zhaojiang Lin \qquad Zihan Liu \qquad Pascale Fung\thanks{$^\star$Equal contributions. This work has been partially funded by ITF/319/16FP and MRP/055/18 of the Innovation Technology Commission, the Hong Kong SAR Government, and School of Engineering Ph.D. Fellowship Award, HKUST, and RDC 1718050-0 of EMOS.AI.}}
\address{
Center for Artificial Intelligence Research (CAiRE)\\
The Hong Kong University of Science and Technology\\
\tt{\{giwinata, scahyawijaya, zlinao, zliucr\}@connect.ust.hk}
}
\begin{document}
%
\maketitle
\begin{abstract}
Highly performing deep neural networks come at the cost of computational complexity that limits their practicality for deployment on portable devices. We propose the low-rank transformer (LRT), a memory-efficient and fast neural architecture that significantly reduces the parameters and boosts the speed of training and inference for end-to-end speech recognition. Our approach reduces the number of parameters of the network by more than 50\% and speeds up the inference time by around 1.35x compared to the baseline transformer model. The experiments show that our LRT model generalizes better and yields lower error rates on both validation and test sets compared to an uncompressed transformer model. The LRT model outperforms those from existing works on several datasets in an end-to-end setting without using an external language model or acoustic data.
\end{abstract}
%

\section{Introduction}
\label{sec:intro}


End-to-end automatic speech recognition (ASR) models have shown great success in replacing traditional hybrid HMM-based models by integrating acoustic, pronunciation, and language models into a single model structure. They rely only on paired acoustic and text data, without additional acoustic knowledge, such as from phone sets and dictionaries. There are two main kinds of end-to-end encoder-decoder ASR architectures. The first is RNN-based sequence-to-sequence (Seq2Seq) models with attention \cite{chan2016listen, kim2017joint}, which learn the alignment between sequences of audio and their corresponding text. The second~\cite{dong2018speech,li2019speechtransformer} applies a fully-attentional feed-forward architecture
transformer~\cite{vaswani2017attention}, which improves on RNN-based ASR in terms of performance and training speed with a multi-head self-attention mechanism and parallel-in-time computation. However, the modeling capacity of both approaches relies on a large number of parameters. Scaling up the model's size increases the computational overhead, which limits its practicality for deployment on portable devices without connectivity and slows down both the training and inference processes.


\begin{figure}[!t]
  \centering
  \includegraphics[width=.72\linewidth]{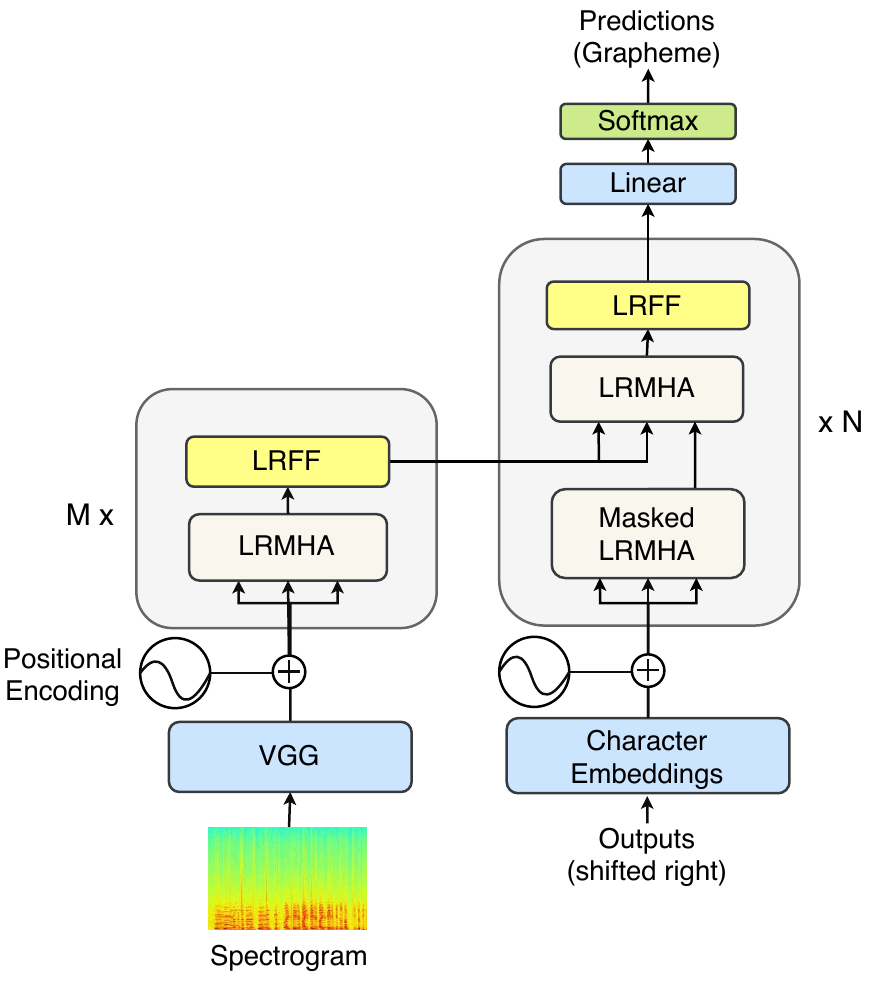}  
  \caption{Low-Rank Transformer Architecture.}
  \label{fig:transformer-asr}
\end{figure}

\begin{figure*}[!t]
\begin{minipage}{.37\textwidth}
  \centering
  \includegraphics[width=.9\linewidth]{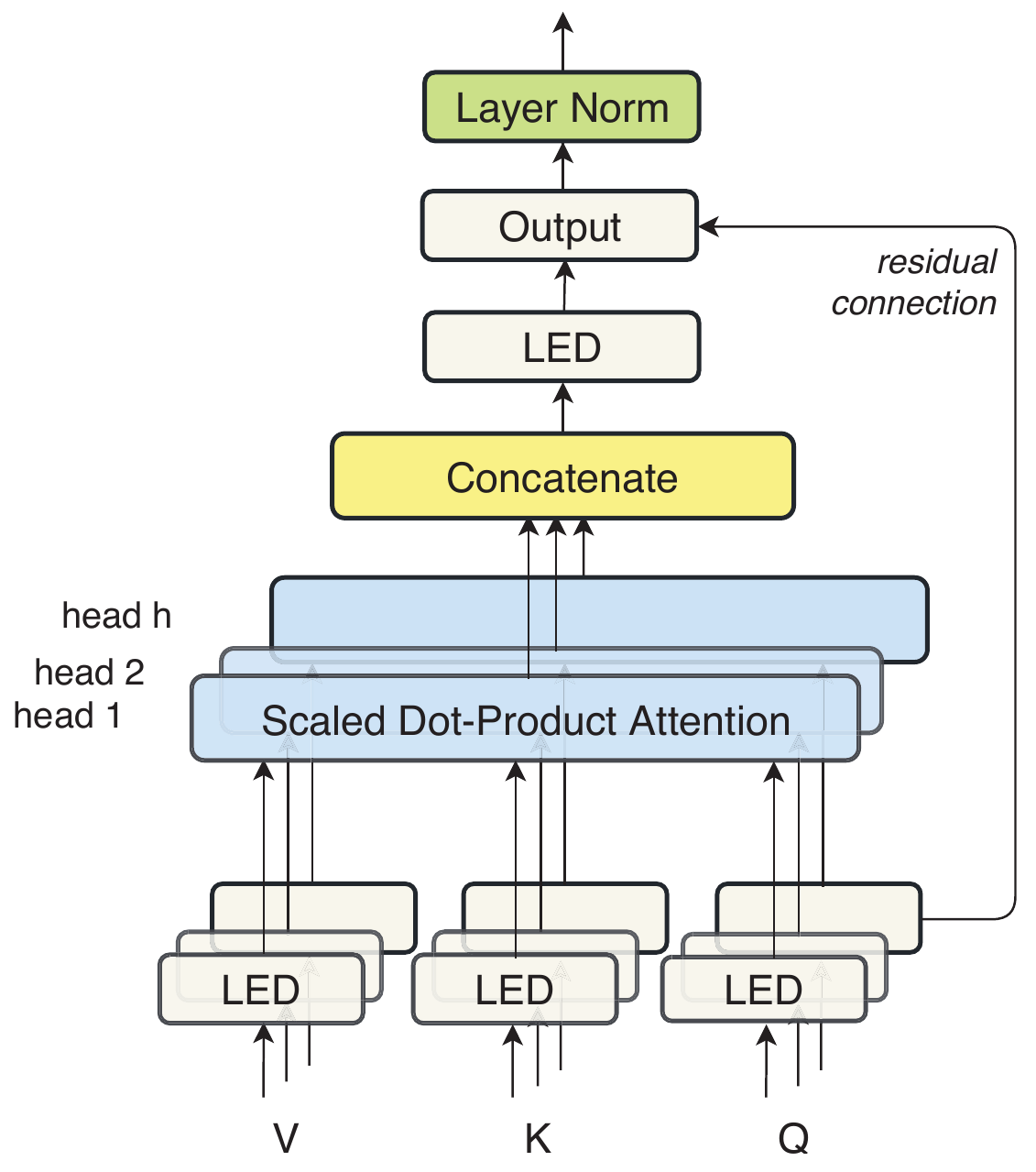}  
\end{minipage}
\begin{minipage}{.15\textwidth}
\end{minipage}
\begin{minipage}{.29\textwidth}
  \centering
  \includegraphics[width=.7\linewidth]{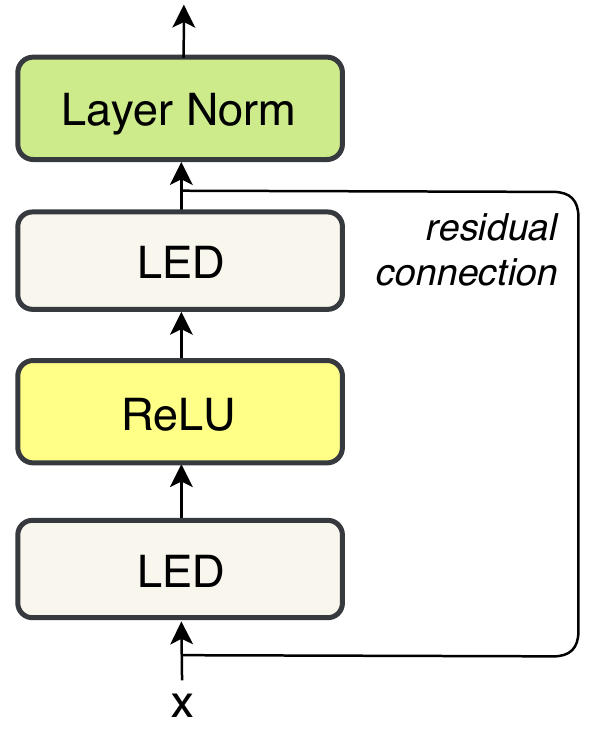}  
\end{minipage}
\begin{minipage}{.25\textwidth}
  \centering
  \includegraphics[width=.57\linewidth]{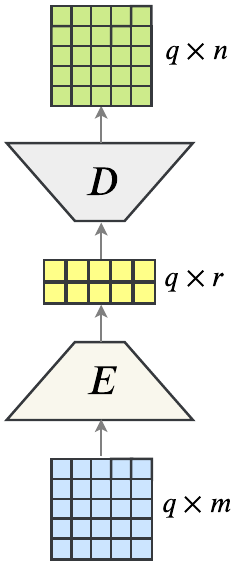}
\end{minipage}
\caption{Low-Rank Transformer Unit. \textbf{Left:} Low-Rank Multi-head Attention (LRMHA), \textbf{Center:} Low-Rank Feed-forward (LRFF), and \textbf{Right:} Linear Encoder-Decoder (LED).}
\label{fig:lr-units}
\end{figure*}


We propose a novel factorized transformer-based model architecture, the low-rank transformer (LRT), to reduce the number of parameters in the transformer model by replacing large high-rank matrices with low-rank matrices to eliminate the computational bottleneck. It optimizes the space and time complexity of the original model when we choose the factorization rank that is relatively smaller than the original matrix dimensions. We design the LRT by taking the idea of the autoencoder that compresses the high-dimensional data input into a compressed vector representation. And, it decodes back to a high-rank matrix to learn latent space representations of the high-rank matrix. This approach is considered an in-training compression method, where we compress the parameters of the model prior to the training process. Our contributions are as follows.
\begin{itemize}
    \item We introduce a novel lightweight transformer architecture leveraging low-rank matrices that outperforms those from existing baselines on the AiShell-1 and HKUST test sets in an end-to-end setting.
    \item We successfully reduce the inference time by up to 1.35x speed-up in the GPU and 1.23x speed-up in the CPU by shrinking the number of parameters by more than 50\% from the baseline.
    \item Interestingly, based on our experiments, we show that our LRT model generalizes better and yield lower error rates on both validation and test performance compared to an uncompressed transformer model.
\end{itemize}

\section{Related Work}
\label{sec:related}

\subsection{Low-Rank Model Compression}
Training end-to-end deep learning ASR models requires high computational resources and long training time to make convergence possible. \cite{sainath2013low} proposed a low-rank matrix factorization of the final weighting layer, which reduced the number of parameters by up to 31\% on a large-vocabulary continuous speech recognition task.
\cite{dudziak2019shrinkml} introduced a reinforcement learning method to compress the ASR model iteratively and learn compression ranks, but it requires more than a week to train. 
In another line of work, a post-training compression method on LSTM using non-negative matrix factorization was proposed by~\cite{winata2019effectiveness} to compress large pre-trained models. However, this technique does not speed-up the training process.
The aforementioned approaches reduce the number of model parameters while keeping the performance loss low. In this work, we extend idea of the in-training compression method proposed in~\cite{kuchaiev2017factorization} by implementing low-rank units on the transformer model~\cite{vaswani2017attention}, which is suitable for effectively shrinking the whole network size and at the same time, reducing the computational cost in training and evaluation, with improvements in the error rate. 

\subsection{End-to-end Speech Recognition}
Current end-to-end automatic speech recognition models are of two main types: (a) CTC-based models~\cite{graves2014towards,audhkhasi2018building}, and (b) Seq2Seq-based models, such as LAS~\cite{chan2016listen}. A combination of both models is also proposed by \cite{hori2017advances}. Recent work by \cite{dong2018speech,li2019speechtransformer} employ a different approach by utilizing the transformer block. The present study is related to these recent transformer ASR approaches, while also leveraging the effectiveness of in-training low-rank compression methods, which was not considered in the aforementioned works.



\begin{table*}[!ht]
\centering
\caption{Results on AiShell-1 (left) and HKUST (right) test sets. For the end-to-end approach, we limit the evaluation to systems without any external data and perturbation to examine the effectiveness of the approach. We approximate the number of parameters based on the description in the previous studies.}
\begin{minipage}{0.49\textwidth}
\centering
\resizebox{0.74\textwidth}{!}{
\begin{tabular}{lcc}
\toprule
\multicolumn{1}{l}{\textbf{Model}} & \multicolumn{1}{l}{\textbf{Params}} & \textbf{CER} \\ \midrule
\multicolumn{3}{c}{\textit{Hybrid approach}} \\ \midrule
\multicolumn{1}{l}{HMM-DNN~\cite{hori2017advances}} & - & 8.5\% \\ \midrule
\multicolumn{3}{c}{\textit{End-to-end approach}} \\ \midrule
\multicolumn{1}{l}{Attention Model~\cite{li2019end}} & - & 23.2\% \\ 
\multicolumn{1}{l}{\hspace{7.4mm} + RNNLM~\cite{li2019end}} & - & 22.0\% \\ 
\multicolumn{1}{l}{CTC~\cite{li2019framewise}} & $\approx$11.7M & 19.43\% \\
\multicolumn{1}{l}{Framewise-RNN~\cite{li2019framewise}} & $\approx$17.1M & 19.38\% \\ 
\multicolumn{1}{l}{ACS + RNNLM~\cite{li2019end}} & $\approx$14.6M & 18.7\% \\ \midrule
\multicolumn{1}{l}{Transformer (large)} & 25.1M & 13.49\% \\
\multicolumn{1}{l}{Transformer (medium)} & 12.7M & 14.47\% \\
\multicolumn{1}{l}{Transformer (small)} & 8.7M & 15.66\%\\ \midrule
\multicolumn{1}{l}{LRT ($r=100$)} & 12.7M & \textbf{13.09\%} \\
\multicolumn{1}{l}{LRT ($r=75$)} & 10.7M & 13.23\% \\
\multicolumn{1}{l}{LRT ($r=50$)} & 8.7M & 13.60\%\\ \bottomrule
\end{tabular}
}
\end{minipage}
\begin{minipage}{0.49\textwidth}
\centering
\resizebox{0.8\textwidth}{!}{
\begin{tabular}{lcc}
\toprule
\multicolumn{1}{l}{\textbf{Model}} & \multicolumn{1}{l}{\textbf{Params}} & \textbf{CER} \\ \midrule
\multicolumn{3}{c}{\textit{Hybrid approach}} \\ \midrule
\multicolumn{1}{l}{DNN-hybrid~\cite{hori2017advances}} & - & 35.9\% \\ 
\multicolumn{1}{l}{LSTM-hybrid (with perturb.)~\cite{hori2017advances}} & - & 33.5\% \\ \midrule
\multirow{2}{*}{\begin{tabular}[c]{@{}l@{}}TDNN-hybrid, lattice-free MMI\\ (with perturb.)~\cite{hori2017advances}\end{tabular}} & \multirow{2}{*}{-} & \multirow{2}{*}{28.2\%} \\ 
& & \\ \midrule
\multicolumn{3}{c}{\textit{End-to-end approach}} \\ \midrule
\multicolumn{1}{l}{Attention Model~\cite{hori2017advances}} & - & 37.8\% \\
\multicolumn{1}{l}{CTC + LM~\cite{miao2016empirical}} & $\approx$12.7M & 34.8\% \\
\multicolumn{1}{l}{MTL + joint dec. (one-pass)~\cite{hori2017advances}} & $\approx$9.6M & 33.9\% \\ 
\multicolumn{1}{l}{\hspace{7.4mm} + RNNLM (joint train)~\cite{hori2017advances}} & $\approx$16.1M & 32.1\% \\ \midrule
\multicolumn{1}{l}{Transformer (large)} & 22M & 29.21\% \\
\multicolumn{1}{l}{Transformer (medium)} & 11.5M & 29.73\% \\
\multicolumn{1}{l}{Transformer (small)} & 7.8M & 31.30\% \\ \midrule
\multicolumn{1}{l}{LRT ($r=100$)} & 11.5M & \textbf{28.95\%} \\
\multicolumn{1}{l}{LRT ($r=75$)} & 9.7M & 29.08\% \\
\multicolumn{1}{l}{LRT ($r=50$)} & 7.8M & 30.74\%\\ \bottomrule
\end{tabular}
}
\end{minipage}
\label{results}
\end{table*}

\section{Low-Rank Transformer ASR}
We propose a compact and more generalized low-rank transformer unit by extending the idea of the in-training compression method proposed in~\cite{kuchaiev2017factorization}. In our transformer architecture, we replace the linear feed-forward unit~\cite{vaswani2017attention} with a factorized linear unit called a linear encoder-decoder (LED) unit. Figure~\ref{fig:transformer-asr} shows the architecture of our proposed low-rank transformer, and Figure~\ref{fig:lr-units} shows the low-rank version of the multi-head attention and position-wise feed-forward network, including the LED. The proposed end-to-end ASR model accepts a spectrogram as the input and produces a sequence of characters as the output similar to~\cite{winata2019code}. It consists of $M$ layers of the encoder and $N$ layers of the decoder. We employ multi-head attention to allow the model to jointly attend to information from different representation subspaces in a different position.


\subsection{Linear Encoder-Decoder}
We propose linear encoder-decoder (LED) units in the transformer model instead of a single linear layer. The design is based on matrix factorization by approximating the matrix $\mathbf{W} \in \mathbb{R}^{m \times n}$ in the linear feed-forward unit using two smaller matrices, $\mathbf{E}  \in \mathbb{R}^{m \times r}$ and $\mathbf{D}  \in \mathbb{R}^{r \times n}$:
\begin{equation}
    \mathbf{W} \approx \mathbf{E} \times \mathbf{D}.
\end{equation}
The matrix $\mathbf{W}$ requires $mn$ parameters and $mn$ flops, while  $\mathbf{E}$ and $\mathbf{D}$ require $rm + rn=r(m+n)$ parameters and $r(m+n)$ flops. If we take the rank to be very low $r<<m,n$, the number of parameters and flops in $\mathbf{E}$ and $\mathbf{D}$ are much smaller compared to $\mathbf{W}$.


\subsection{Low-Rank Multi-Head Attention}
The LED is incorporated into the multi-head attention by factorizing the projection layers of keys $W_i^Q$, values $W_i^V$, queries $W_i^Q$, and the output layer $W^O$. A residual connection from a query $Q$ to the output is added.
\begin{align}
&\textnormal{Attention}(Q,K,V)=\textnormal{Softmax}(\frac{QK^T}{\sqrt{d_k}}V),\\
&hd_i = \textnormal{Attention}(QE_i^Q D_i^Q, KE_i^K D_i^K, VE_i^V D_i^V),\\
&f(Q,K,V) = \textnormal{Concat}(h_1,\cdots,h_H)E^O D^O+Q,
\end{align}
where $f$ is a low-rank multi-head attention (LRMHA) function, $h_i$ is the head of $i$, $H$ is the number of heads, and the projections are parameter matrices $E_i^Q \in \mathbb{R}^{d_{model} \times d_r}$; $D_i^Q \in \mathbb{R}^{d_{r} \times d_k}$; $E_i^K \in \mathbb{R}^{d_{model} \times d_r}$; $D_i^K \in \mathbb{R}^{d_{r} \times d_k}$; $E_i^V \in \mathbb{R}^{d_{model} \times d_r}$; $D_i^V \in \mathbb{R}^{d_{r} \times d_v}$. $d_{model}$, $d_{r}$, $d_{k}$, and $d_{v}$ are dimensions of hidden size, rank, key, and value, respectively.
\subsection{Low-Rank Feed-Forward}
Each encoder and decoder layer has a position-wise feed-forward network that contains two low-rank LED units and applies a ReLU function in between. To alleviate the gradient vanishing issue, a residual connection is added, as shown in Figure \ref{fig:lr-units}.
\begin{equation}
    g(x) = \textnormal{LayerNorm}(  \textnormal{max}(0, x E_1 D_1)E_2 D_2 + x),
\end{equation}
where $g$ is a low-rank feed-forward (LRFF) function.


\subsection{Training Phase}
The encoder module uses a VGG net~\cite{simonyan2014very} with a 6-layer CNN architecture. The VGG consists of convolutional layers that are added to learn a universal audio representation and generate input embedding. The input of the unit is a spectrogram. The decoder receives the encoder outputs and applies multi-head attention to the decoder input. We apply a mask in the attention layer to avoid any information flow from future tokens. Then, we run a non-autoregressive step and calculate the cross-entropy loss.


\subsection{Evaluation Phase}
In the inference time, we decode the sequence using autoregressive beam-search by selecting the best sub-sequence scored using the softmax probability of the characters. We define $P(Y)$ as the probability of the sentence. A word count is added to avoid generating very short sentences. $P(Y)$ is calculated as follows:

\begin{equation}
P(Y) = \alpha P_{trans}(Y|X) + \gamma \sqrt{wc(Y)},
\end{equation}

\noindent where $\alpha$ is the parameter to control the decoding probability from the decoder $P_{trans}(Y|X)$, and $\gamma$ is the parameter to control the effect of the word count $wc(Y)$.

\section{Experiments}

Experiments were conducted on two dataset benchmarks: AiShell-1~\cite{bu2017aishell}, a multi-accent Mandarin speech dataset, and HKUST~\cite{liu2006hkust}, a conversational telephone speech recognition dataset. The former consists of 150 hours, 10 hours, and 5 hours of training, validation, and testing, respectively, while the latter consists of a 5 hour test set, 4.2 hours extracted from the training data as the validation set, and the remaining 152 hours as the training set.
We concatenate all characters in the corpus, including three special tokens, such as $<$PAD$>$, $<$SOS$>$, and $<$EOS$>$. In our models, we use two encoder layers and four decoder layers. The large transformer consists of a $dim_{inner}$ of 2048, $dim_{model}$ of 512, and $dim_{emb}$ of 512. For the smaller transformers, we select the same parameters as the LRT model with $r=100$, $r=75$ and $r=50$. In the beam-search decoding, we take $\alpha = 1$, $\gamma = 0.1$, and a beam size of 8. We evaluate our model using a single GeForce GTX 1080Ti GPU and three Intel Xeon E5-2620 v4 CPU cores. We use character error rate (CER) as the evaluation metric.

\section{RESULTS AND DISCUSSION}
\subsection{Evaluation Performance}
\begin{figure}[!t]
\centering
\includegraphics[width=\linewidth,height=6.2cm]{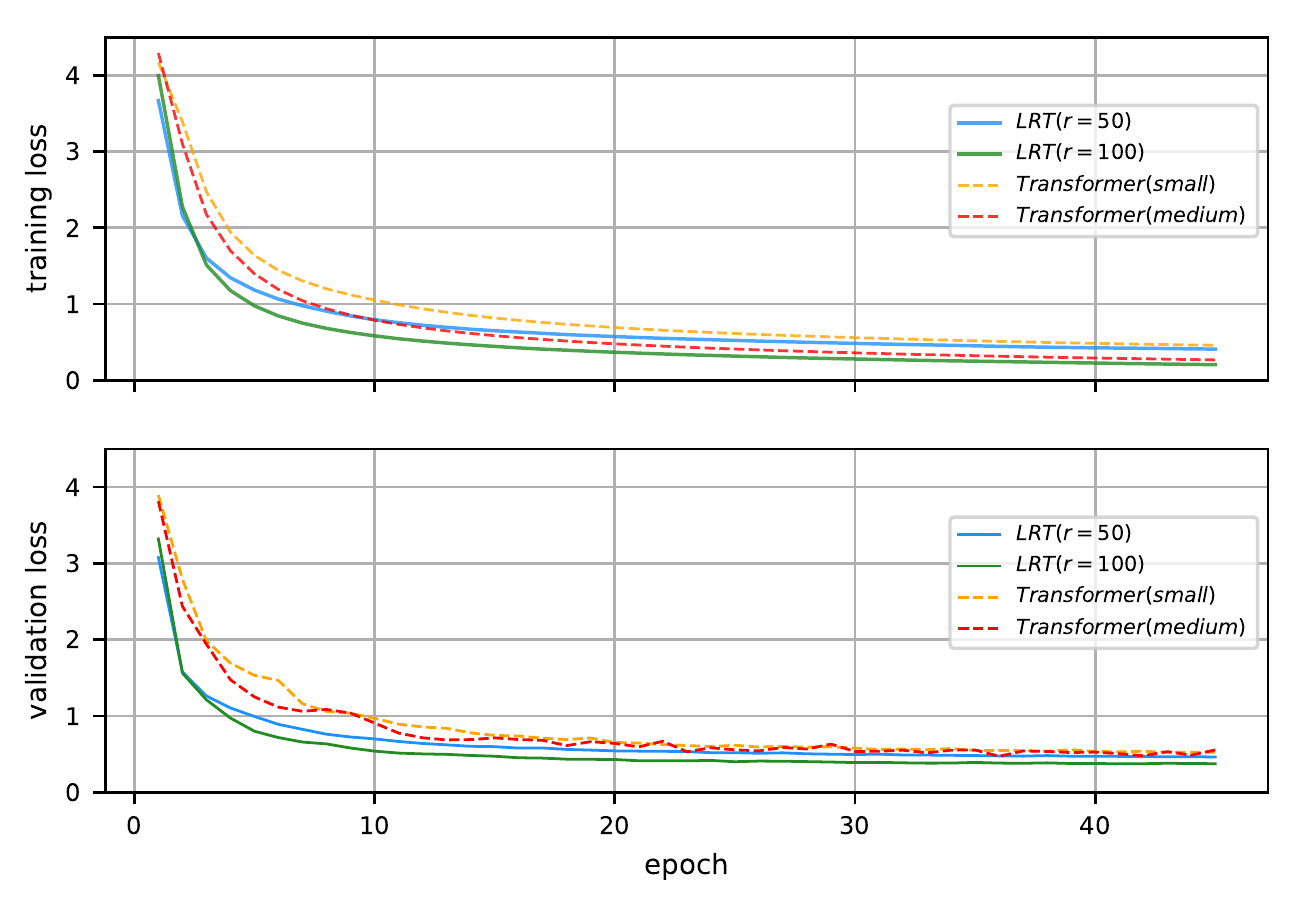}
\caption{Training and validation losses on AiShell-1 data.}
\label{fig:loss-trend}
\end{figure}

Table~\ref{results} shows the experiment results. LRT models gain slight improvement even with a greater than a 50\% compression rate, and they outperform the vanilla transformers on both the AiShell-1 and HKUST test sets, with a 13.09\% CER and a 28.95\% CER, respectively. In addition, we further minimize the gap between the HMM-based hybrid and end-to-end approaches without leveraging a perturbation strategy or external language model. Interestingly, our LRT models achieve lower validation loss compared to the uncompressed Transformer (large) baseline model, which implies that our LRT models regularize better, as shown in Figure~\ref{fig:loss-trend}. The models are faster to converge and stop in a better local minimum compared to the vanilla transformers.

\subsection{Memory and Time Efficiency}
As shown in Table~\ref{results}, our LRT ($r=50$) model achieves similar performance to the large transformer model despite having only one-third of the large transformer parameters. In terms of time efficiency, our LRT models gain inference time speed-up by up to 1.35x in the GPU and 1.23x in the CPU, and 1.10x training time speed-up in the GPU compared to the uncompressed Transformer (large) baseline model, as shown in Table \ref{efficiency}. We also compute the average length of the generated sequences to get a precise comparison. In general, both the LRT and baseline models generate sequences with a similar length, which implies that our speed-up scores are valid.

\begin{table}[!t]
\centering
\caption{Compression rate and inference speed-up of LRT models vs. Transformer (large). $\Delta$CER and $|\bar{X}|$ denote the improvement, and the mean length of generated sequences.}
\resizebox{0.475\textwidth}{!}{
\begin{tabular}{lccccccc}
\toprule
\multicolumn{1}{c}{\multirow{2}{*}{\textbf{dataset}}} & \multirow{2}{*}{\textbf{r}} & \multirow{2}{*}{\textbf{$\Delta$CER}} & \multicolumn{1}{c}{\multirow{2}{*}{\textbf{compress.}}} & \multicolumn{2}{c}{\textbf{speed-up}} & \multicolumn{1}{c}{\multirow{2}{*}{\textbf{$|\bar{X}|$}}} \\ \cmidrule{5-6}
\multicolumn{1}{c}{} & & & & \multicolumn{1}{c}{\textbf{GPU}} & \textbf{CPU only} & \\ \midrule
AiShell-1 & base & 0 & 0 & 1 & 1 & 23.08 \\
 & 100 & 0.40\% & 49.40\% &  1.17x & 1.15x & 23.15 \\
 & 75 & 0.26\% & 57.37\% & 1.23x & 1.16x & 23.17 \\
 & 50 & -1.10\% & 65.34\% & 1.30x & 1.23x & 23.19 \\ \midrule
HKUST & base & 0 & 0 & 1 & 1 & 22.43 \\
 & 100 & 0.26\% & 47.72\% & 1.21x & 1.14x & 22.32 \\
 & 75 & 0.13\% & 55.90\% & 1.26x & 1.15x & 22.15 \\
 & 50 & -1.53\% & 64.54\% & 1.35x & 1.22x & 22.49 \\ \bottomrule
\end{tabular}
}
\label{efficiency}
\end{table}

\section{Conclusion}
We propose low-rank transformer (LRT), a memory-efficient and fast neural architecture that compress the network parameters and boosts the speed of the inference time by up to 1.35x in the GPU and 1.23x in the CPU, as well as the speed of the training time for end-to-end speech recognition. Our LRT improves the performance even though the number of parameters is reduced by 50\% compared to the baseline transformer model. Our approach generalizes better than uncompressed vanilla transformers and outperforms those from existing baselines on the AiShell-1 and HKUST datasets in an end-to-end setting without using additional external data.

\vfill\pagebreak
\bibliographystyle{IEEEbib}
\bibliography{strings,refs}

\end{document}